\documentclass[conference]{IEEEtran}
\usepackage{times}

\usepackage[numbers]{natbib}
\usepackage{multicol}
\usepackage[bookmarks=true,linkcolor={red!50!black}]{hyperref}

\usepackage{amsmath, amssymb, stmaryrd}
\usepackage[table,xcdraw,dvipsnames]{xcolor}
\usepackage{graphicx}
\usepackage{booktabs}
\usepackage{subcaption}
\usepackage{microtype}
\usepackage[normalem]{ulem}
\usepackage{adjustbox}
\usepackage{multirow}
\usepackage{tabularx}
\usepackage{makecell}
\usepackage{etoc}
\usepackage[accsupp]{axessibility}
\usepackage{xspace}
\usepackage{diagbox}
\usepackage{wrapfig}

\usepackage{bbding}
\usepackage{pifont}
\usepackage{wasysym}

\usepackage{siunitx}

\usepackage{url}
\usepackage{cleveref} 
\crefname{section}{Sec.}{Secs.}
\Crefname{table}{Tab.}{Tabs.}
\Crefname{figure}{Fig.}{Figs.}
\crefname{equation}{Eq.}{Eqs.}
\crefname{appendix}{Apx.}{Apx.}

\newcommand{\gain}[1]{\textcolor{Green}{+{#1}}}

\usepackage{pifont}
\usepackage{bbding}

\usepackage{crossreftools}

\usepackage{cuted}

\definecolor{citecolor}{HTML}{0071BC}
\hypersetup{colorlinks=true,citecolor=RoyalBlue}

\crefname{section}{Sec.}{Secs.}
\Crefname{section}{Sec.}{Secs.}
\Crefname{table}{Tab.}{Tabs.}
\crefname{table}{Tab.}{Tabs.}
\Crefname{figure}{Fig.}{Figs.}
\crefname{figure}{Fig.}{Figs.}

\newcommand\clearrow{\global\let\rowmac\relax}

\usepackage{amsmath,amsfonts,bm}

\def\eqref#1{equation~\ref{#1}}

\def\1{\bm{1}}

\DeclareMathAlphabet{\mathsfit}{\encodingdefault}{\sfdefault}{m}{sl}
\SetMathAlphabet{\mathsfit}{bold}{\encodingdefault}{\sfdefault}{bx}{n}

\usepackage{ifthen}
\newboolean{review}
\setboolean{review}{false} 

\ifthenelse{\boolean{review}}{

\usepackage{ulem}
\usepackage{marginnote}

\newcommand{\OB}[1]{\textbf{Comment:#1}}
\newcommand{\REP}[1]{\textbf{Reply:#1}}

\newcommand{\fix}[1]{{\color{red}#1}}
\newcommand{\fixs}[1]{{\color{red}\sout{#1}}}

\newcommand{\fixS}[2]{\fixs{#1\protect\marginnote{\color{red}#2}\label{#2}}}
\newcommand{\fixM}[2]{\fix{\protect\marginnote{\color{red}#2}#1\label{#2}}}
\newcommand{\fixN}[2]{\fix{#1\label{#2}\protect\marginnote{\color{red}#2}}}

\newcommand{\cancel}[1]{\fix{#1}}

\newcommand{\fixRef}[1]{\textcolor{blue}{\bf #1 on \cpageref{#1}}}

}{

\newcommand{\fix}[1]{#1}
\newcommand{\fixS}[2]{}
\newcommand{\fixs}[1]{}
\newcommand{\fixM}[2]{#1}
\newcommand{\fixN}[2]{#1}
\newcommand{\cancel}[1]{}

}

\begin{document}

\ifthenelse{\boolean{review}}{
\thispagestyle{plain}
\pagestyle{plain}
}{
\pagenumbering{gobble}
}

\ifthenelse{\boolean{review}}{
\pagenumbering{roman}
\clearpage
\newpage
\section*{Author Response to Reviews}

We would like to thank all reviewers for their insightful review and constructive comment.
We have addressed all concerns in the revised version of the paper and summarised them here.
Alongside the letter, we have provided an annotated version in which all changes have been highlighted and marked with a \fixM{unique ID indicated in the margins}{1.1 (e.g.)} to help the reviewers locate the portion of the work that addresses their concerns.
\textcolor{blue}{\bf Blue text in this letter} maps to any changes made to the manuscript. 

\subsection*{Reviewer \#1 (tEV2)}
\noindent\OB{1.1} \textit{Auto-regressive models are known to perform not great for predicting floating numbers. Have the authors explored using other representation for action (speed and course numbers)
}\\
\REP{1.1} Thank you for raising this point. While it is true that general-purpose Large Language Models (LLMs) are known for their non-ideal performance in predicting floating-point numbers \citep{golkar2023xval}, our approach involves fine-tuning the numerical language tokens with a paired visual-language-action dataset, known as symbol tuning \cite{wei2023symbol}. The effectiveness of such approach, where instead of outputting floats directly an action signal is represented as floating-point numbers by tokenizing the string of digits, has also been demonstrated in several related works, including manipulation (e.g., RT-2 \cite{brohan2023rt}, RT-X \cite{padalkar2023open}) and driving applications (e.g., DriveGPT4 \cite{xu2023drivegpt4}). Therefore, in our study, we adopted this approach due to its validated effectiveness in this related research and its intuitive alignment with human understanding. We consider exploring alternative methods for action representation an promising avenue for future research. 
We have included a further discussion on this topic in \cref{appx:action} (\fixRef{1.1}).
\\\\
\OB{1.2} \textit{For the pretraining stage, the authors mainly used generic video-caption pairs. Would it make sense to use a domain specific pre training dataset that has more driving data.}
\\
\REP{1.2} This is a good point. We agree on the hypothesis that using more domain specific dataset for pre-training can be beneficial. 
However, our choice of a generic video-caption dataset for pre-training is for two major reasons: 
\textbf{(1)}  The purpose of the pre-training stage is to align the visual representation from the video encoder with the language representation in the LLM backbone.
Using a web-scale, generic video-caption dataset means that we learn this projection more robustly, as this covers a broader range of visual and language features.
This approach is also aligned with the paradigm \cite{liu2023llava} of pre-training a general VLM backbone on large-scale data and then fine-tuning it with a relatively small amount of domain- or task-specific  data for specialisation.
\textbf{(2)} Practically, the scale of any available driving-specific video-caption dataset is considerably smaller than its generic counterparts.
By the time of our initial writing, driving-language datasets consist mostly of driving-explanation pairs (e.g. BDD-X \cite{kim2018bddx}), which is not suitable for pre-training the language-video alignment in terms of scale and language format.
While there are more recent, emerging datasets such as DriveMLM \cite{wang2023drivemlm} and Reason2Drive \cite{nie2023reason2drive} which curate larger scale and more annotation on this manner providing promising future potential, we note that by the time of our research these datasets are not yet open-sourced (e.g. no release at \url{github.com/OpenGVLab/DriveMLM}, to date and only ``mini-version'' released at \url{github.com/fudan-zvg/Reason2Drive}).
We have included notes to this effect in the manuscript in \cref{sec:datascarce} (\fixRef{1.2}).
\\\\
\OB{1.3} \textit{In Fig.1 I.D performance and OOD performance has different scale in y axis, this seems confusing and misleading.}
\\
\REP{1.3} Thank you for raising this point. The different scale  in \cref{fig:front} is mainly for the purpose of visualisation, since difference between I.D and OOD performance is large and this makes the comparison in OOD performance less visually distinguishable if plotted on the same scale. However, to better guide the reader, we have modified both the caption of \cref{fig:front} (\fixRef{1.3.1}) and reference to it in the text (\fixRef{1.3.2}).
\\\\
\OB{1.4} \textit{The retrieval part mostly samples 2 examples from the database, is this because of the context length of the model? Have the authors studied the effect of different numbers of samples to be added as ICL examples?}
\\
\REP{1.4} Yes, the design choice of $2$ ICL examples is due to the restriction in the context window of the backbone LLM. Specifically, the context window of our LLM backbone, Vicuna 1.5 \cite{zheng2023judging}, is $4096$ tokens, and each additional ICL example introduces roughly an extra $1800$ tokens, including a $1024$ fixed-length video token sequence and remaining tokens of language dialogue. 
This means we cannot exceed $1800\times2=3600<4096$ tokens corresponding to two ICL examples, as e.g. $1800\times3=4400>4096$.
The only way to include more examples would be to reduce the token length of the video examples, but this is fixed by the MLLM architecture that we have chosen, Video-LLaVA~\cite{lin2023video}.
However, we have now included an additional ablation study on the effect of varying the number of in-context learning examples from $1$ to $2$.
We include these further experiment results and analysis in \cref{sec:generalisation} (\fixRef{1.4.1}) and \cref{sec:comp} (\fixRef{1.4.2}).

\subsection*{Reviewer \#2 (6kSH)}
\noindent\OB{2.1} \textit{Fig. 1 is a bit difficult to understand without first reading the paper (the right side plot in particular). It is unclear what are the metrics and the baselines.}
\\
\REP{2.1} Thank you for raising this point. We have modified the caption and text labels of \cref{fig:front} (\fixRef{2.1}) to better guide the reader to understand the visualised metrics and baselines.
\\\\
\OB{2.2} \textit{In Fig. 2, it is unclear what the MLP projections do from looking at the figure or reading the caption.}
\\
\REP{2.2} We have modified the caption of \cref{fig:main} (\fixRef{2.2}) to better highlight the role of MLP projections. This comment has also incidentally alerted us to a problem with figure numbering, where Fig. 2 was skipped (numbering went from Fig. 1 to Fig. 3), we have amended this and all Figure numbers from 3 onwards have been decremented.
\\\\
\OB{2.3} \textit{In \cref{sec:training}, it was a bit strange to refer to information in future sections.}
\\
\REP{2.3} We have modified the related part in \cref{sec:training} (\fixRef{2.3.1}, \fixRef{2.3.2}) to avoid referring to future sections.
\\\\
\OB{2.4} \textit{In Fig. 5, it is not clear to me how the video is included in the in-context prompt.}
\\
\REP{2.4} We have modified the caption of \cref{fig:sample} (\fixRef{2.4}, note figure numbering change as per \OB{2.2} above) to highlight the way video tokens are incorporated in ICL examples.
\\\\
\OB{2.5} \textit{In Fig. 3, row 3, column 2, is the response a hallucination?}
\\
\REP{2.5} Thank you for raising this point. 
By observing the generated answers and ground truth annotation, we agree the mentioned case is a hallucination. We hypothesise
that this is due to the limited capacity of small-scale models
which rely on spurious correlations in limited training data, as the model associates the car’s action of slowing down with the presence of a stop
sign, which does not exist by observation.
We have included a brief discussion of this in \cref{sec:hallucinate} (\fixRef{2.5}), being a well known issue with LLMs.
\\\\
\OB{2.6} \textit{The generalist vs specialist approach categories should be described in the experiments sections.}
\\
\REP{2.6} Thank you for raising this point. We have included additional description about the difference between generalist and specialist models in \cref{sec:settings_datasets} (\fixRef{2.6}).
\\\\
\OB{2.7} \textit{The following is a non-exhaustive list of typos in the paper:
\begin{itemize}
\item 2.7.1: Fig. 1 caption: 'inferring' -\> 'infers',
\item 2.7.2: 'specified [in] Sec. IV',
\item 2.7.3: 'significant margin driving explanation'.
\item 2.7.4: Sec. II: 'which help to [decode]'.
\item 2.7.5: Sec. III.A: '[an] MLP', 'both [the] textual action explanation and [the] numerical control signal'.
\item 2.7.6: Equations should part of the sentence (period after the equation if no further explanation for it).
\item 2.7.7: Fig. 4: '[a language-aligned] video representation'.
\item Sec. III.B: 'ablation study [in] Sec. IV.D'.
\item 2.7.8: Sec. III.C: 'In particular...' is a run-on sentence.
\item 2.7.9: Fig. 5: 'Why does the ego car [do] this?'.
\item 2.7.10: Sec. IV.A: 'Spoken-London [dataset]', 'Implementation [Details]', 'Moreover, for the'.
\item 2.7.11: Fig. 6 caption: '[Spoken-London]'.
\item 2.7.12: Sec. IV.F: '[qualitative] examples'.
\item 2.7.13: References should be consistent in format and currently have a number of typos (e.g., year and 'IEEE' is listed in two different places in a citation, conferences are not consistently capitalized and formatted, etc.).
\end{itemize}}
\noindent\REP{2.7} We have fixed these typos, please see:
\fixRef{2.7.1}, \fixRef{2.7.2}, \fixRef{2.7.3}, \fixRef{2.7.4}, \fixRef{2.7.5.1} and \fixRef{2.7.5.2}, \fixRef{2.7.6}, \fixRef{2.7.7}, \fixRef{2.7.8}, \textcolor{blue}{\bf 2.7.9 (graphic corrected in \cref{fig:sample} on \cpageref{fig:sample})}, \fixRef{2.7.10.1}, \fixRef{2.7.10.2}, and \fixRef{2.7.10.3}, \fixRef{2.7.11}, \fixRef{2.7.12}.
\\\\
We have improved correctness and consistency for the bibliography throughout, from \fixRef{2.7.13.1} to \fixRef{2.7.13.50}.
\\\\
We have also found several other typos, fixed at:
\fixRef{2.7.14}, \fixRef{2.7.15}, \fixRef{2.7.16}, \fixRef{2.7.17}, \fixRef{2.7.18}, \fixRef{2.7.19}, \fixRef{2.7.20}, \fixRef{2.7.21}, \fixRef{2.7.22}, \fixRef{2.7.23}, \fixRef{2.7.24}, \fixRef{2.7.25}, \fixRef{2.7.26}, \fixRef{2.7.27}, \fixRef{2.7.28}, \fixRef{2.7.29}, \fixRef{2.7.30}, \fixRef{2.7.31}, \fixRef{2.7.32}, \fixRef{2.7.33}, \fixRef{2.7.34}, \fixRef{2.7.35}, \fixRef{2.7.36}, \fixRef{2.7.37}, \fixRef{2.7.38}, \fixRef{2.7.39}, \fixRef{2.7.40}, \fixRef{2.7.41}, \fixRef{2.7.42}, \fixRef{2.7.43}, \fixRef{2.7.44}, \fixRef{2.7.45}, \fixRef{2.7.46}, \fixRef{2.7.47}, \fixRef{2.7.48}, \fixRef{2.7.49}, \fixRef{2.7.50}, \fixRef{2.7.51}, \fixRef{2.7.52}, \fixRef{2.7.53}.

\subsection*{Reviewer \#3 (FG3o)}

\noindent\OB{3.1} \textit{It would be useful to include more experiment results. In particular, it would be interesting to see whether fine-tuning/training the retrieval engine on the Spoken-London dataset yields improved performance.}
\\
\REP{3.1} Thank you for your suggestion. We have conducted additional experiments to fine-tune the retrieval engine on the Spoken-London dataset. While RAG-Driver aims to enhance the model's generalisability in driving action explanation tasks in a training-free manner through in-context learning, we observe that training retrieval engine with data samples from the target domain can improve performance in particular in O.O.D settings, indicating the potential of further learning-based optimisation of retrieval strategy. We included the experiment results and further discussion in \cref{sec:fine-tuning-experiments} \fixRef{3.1}.
\\\\
\OB{3.2} \textit{Since this is a robotics conference, it would be good to evaluate action prediction in a close-loop fashion using simulations. For example, DriveMLM evaluates close-loop driving using CARLA. This experiment would provide the additional benefit of evaluating this method in another new domain.}
\\
\REP{3.2} Thank you for your suggestion. We agree that closed-loop evaluation in the simulator is beneficial and impactful. However, the main obstacle to apply our MLLM-based driving agent under closed-loop simulation is data scarcity and a significant simulation-to-real gap. Specifically, our method is trained with the real-world dataset BDD-X \cite{kim2018bddx}, unlike DriveMLM \cite{wang2023drivemlm}, which collects a large-scale simulation dataset in CARLA and tests it in the same environment for evaluation. In contrast, we train on the real dataset of BDD-X and test on the out-of-distribution (OOD) real dataset Spoken-London as a zero-shot generalisation, which is closer to the deployment requirements in applications.
While our method mitigates the effects of distribution shift between real datasets collected in different locations and illumination conditions, as indicated in \cref{fig:front}, the larger simulation-to-real gap still remains a significant challenge for a closed-loop testing.
In principle, our method can also be trained with the simulator-collected dataset from DriveMLM, which we expected would circumvent the issue with the simulation-to-real gap. We note, however, that this dataset is not open-sourced yet.
This in turn highlights the importance of developing a generalisable model for the deployment of the MLLM-based driving agent, as part of our early exploration conclusion. We included further discussion in \cref{sec:closedloop} (\fixRef{3.2}).
\\\\
\OB{3.3} \textit{Why the action prediction results for the first two ablations are N/A?}
\\
\REP{3.3} Thank you for raising this point. The N/A values for control signal predictions indicate that, when ICL examples are not provided during training but solely during inference, the MLLM is unable to perform predictions; as the model generates non-floating-point numbers, which precludes the calculation of Mean Square Error (MSE) for quantifying performance. We have modified \cref{sec:ablation} accordingly \fixRef{3.3} to clarify this.
\\\\
\OB{3.4} \textit{Does the Spoken-London dataset include action annotations? If it does, why are the models not evaluated on action predictions?}
\\
\REP{3.4} 
In this work, we primarily focus on investigating the driving action explanation and justification capabilities of RAG-Driver, particularly with respect to generalisation under distribution shifts. We evaluate the explanation performance on the Spoken-London dataset in a zero-shot manner. However, due to the partial annotation of the control signal, we did not evaluate our model on control action signal predictions.
\\\\
\OB{3.5} \textit{What are the training time and inference speed?}
\\
\REP{3.5} Thank you for raising this point. Training of the retrieval engine on the BDD-X dataset using a single A100 GPU takes half an hour. Then, we utilize the pre-training checkpoint from Video-LLaVA \cite{lin2023video} and fine-tune the MLLM on top of it, which takes 6 hours on 8 A100 GPUs. For the inference time, since we store all driving experience in a preprocessed database, which speeds up the retrieval process, the overall single round inference time is roughly 4 seconds on a single A100 GPU. We have included these in \cref{sec:settings_datasets} \fixRef{3.5}.
\\\\
\OB{3.6} \textit{Are the control signals c for retrieval from previous time steps?}
\\
\REP{3.6} Yes, the control signals c for retrieval is provided in the same format as current query, which is the action from previous time steps.
\\\\
\OB{3.7} \textit{What are a, p, and n in Section III.C?}
\\
\REP{3.7} We use a metric learning scheme for training the retrieval engine, where we define the anchor sample ($\bm{a}$), positive sample ($\bm{p}$), and negative sample ($\bm{n}$) and calculate the loss as in \cref{eq:tripletmargin}. This is to ensure the samples driving video with similar driving explanation answers are close together in the metric space.
We have modified the related part in \cref{sec:RAGICL} \fixRef{3.7} to clarify this.
\clearpage
\newpage
}{}

\title{\huge \bf
\textit{RAG-Driver}: Generalisable Driving Explanations with Retrieval-Augmented In-Context Learning in Multi-Modal Large Language Model 
}



\author{
Jianhao Yuan$^1$, 
Shuyang Sun$^1$, 
Daniel Omeiza$^1$, 
Bo Zhao$^2$, 
Paul Newman$^1$, 
Lars Kunze$^1$, 
Matthew Gadd$^1$\\
$^1$ University of Oxford
$^2$Beijing Academy of Artificial Intelligence\\
\texttt{\{jianhaoyuan,kevinsun,daniel,pnewman,lars,mattgadd\}@robots.ox.ac.uk}
}

\makeatletter
\let\@oldmaketitle\@maketitle
\renewcommand{\@maketitle}{\@oldmaketitle
\centering
\includegraphics[width=.95\linewidth]{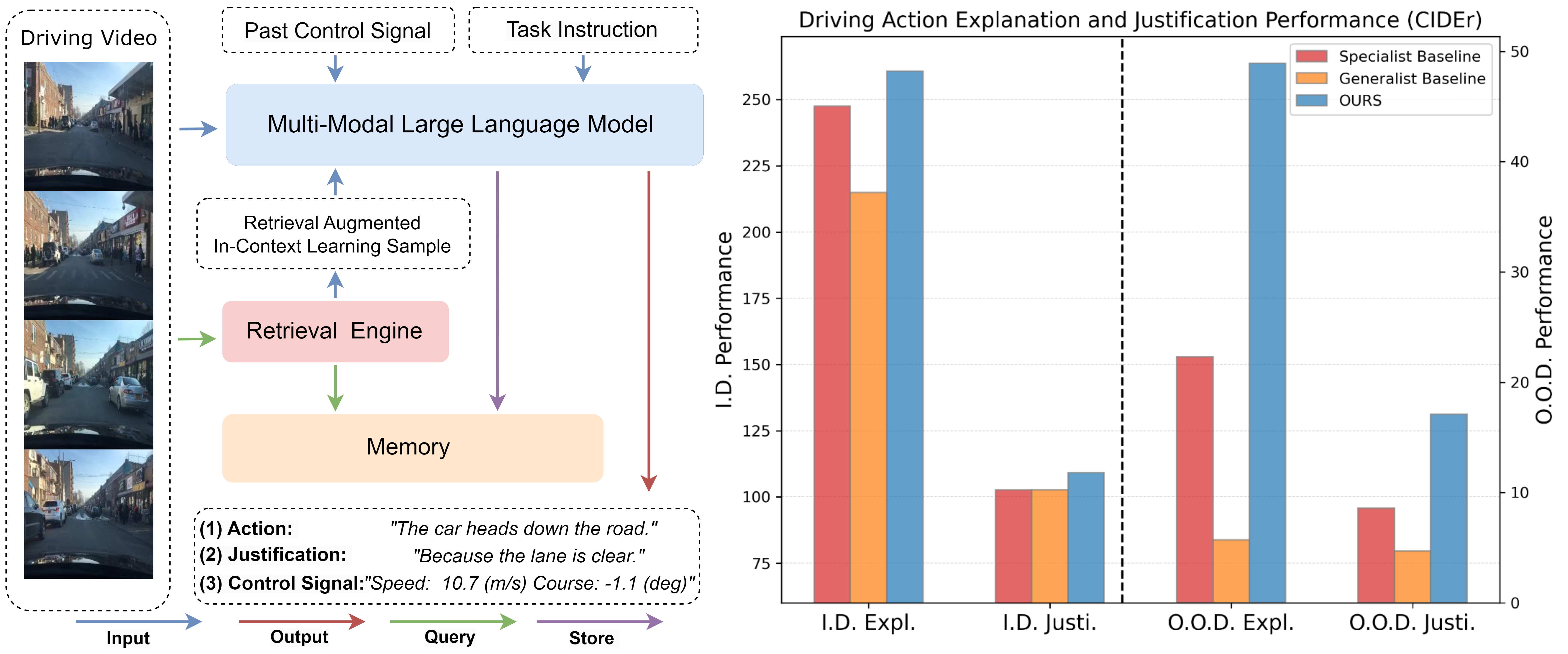}
\captionof{figure}{\label{fig:front} \textbf{Left:} In natural language, our system describes \textit{and} justifies actions taken by the vehicle, as well as \fixM{infers}{2.7.1} the driving action in the form of numerical control signals (speed, and steering angle).
For this, we use a unified perception and planning module through a Multi-modal Large Language Model.
Our core contribution is a \textit{retrieval mechanism} to search driving scenarios similar to the current condition and use these to augment the current predictions through In-Context Learning.
This leads to better overall description and prediction and is more generalisable in new deployment domains. \textbf{Right:} \fixM{Comparison between our methods and baselines for the in-distribution (I.D, left vertical axis) and out-of-distribution (O.O.D, right vertical axis) generalisation settings as specified in~\cref{sec:experiments}.}{2.7.2}
\fixM{Note that I.D. Performance and O.O.D. Performance have different vertical scales.}{1.3.1}
\fixM{Despite the performance gap between these two settings our method significantly outperforms all of the baselines, including the specialist baseline, ADAPT \cite{jin2023adapt}, as well as the generalist baseline, DriveGPT4 \cite{xu2023drivegpt4}, in terms of driving action explanation (Expl. on the horizontal axis) and justification (Justi. on the horizontal axis)\fixM{}{2.1} tasks, as measured by CIDEr.}{2.7.3}. 
}
}
\makeatother

\maketitle

\begin{abstract}
%
%



We need to trust robots that use often opaque AI methods. They need to explain themselves to us, and we need to trust their explanation. In this regard, explainability plays a critical role in trustworthy autonomous decision-making to foster transparency and acceptance among end users, especially in complex autonomous driving.
Recent advancements in Multi-Modal Large Language models (MLLMs) have shown promising potential in enhancing the explainability as a driving agent by producing control predictions along with natural language explanations. 
However, severe data scarcity due to expensive annotation costs and significant domain gaps between different datasets makes the development of a robust and generalisable system an extremely challenging task. 
Moreover, the prohibitively expensive training requirements of MLLM and the unsolved problem of catastrophic forgetting further limit their generalisability post-deployment.
To address these challenges, we present \textit{RAG-Driver}, a novel retrieval-augmented multi-modal large language model that leverages in-context learning for high-performance, explainable, and generalisable autonomous driving.
By grounding in retrieved expert demonstration, we empirically validate that \textit{RAG-Driver} achieves state-of-the-art performance in producing driving action explanations, justifications, and control signal prediction. 
More importantly, it exhibits exceptional zero-shot generalisation capabilities to unseen environments without further training endeavours.

\end{abstract}
\begin{IEEEkeywords}
Autonomous driving, multi-modal language model, end-to-end driving, domain generalisation
\end{IEEEkeywords}

\IEEEpeerreviewmaketitle

\section{Introduction}
Driven by the emerging development of deep learning, autonomous driving has observed a paradigm shift from rules-based decision systems \cite{schwarting2018planning, faisal2019understanding} to data-driven learning-based approaches \cite{Hu_2023_CVPR, bojarski2016end, kendall2019learning}. 
However, this comes at the cost of transparency in decision-making, especially for end-to-end autonomous driving systems which are considered black-box in nature \cite{chen2023end}. Thus, in addition to precision in action control, explanation provision is key \fixM{to}{2.7.15} ensuring trustworthy decision-making to reconcile the system's decisions with end-user expectations to foster confidence and acceptance \cite{yang2023survey, bonnefon2016social} in dynamic driving environments.

Traditional approaches have mainly relied on attention visualisation \cite{bojarski2016visualbackprop, bojarski2017explaining,moriVisualExpl} as a proxy to rationalise the decisions of the black-box systems or auxiliary intermediate tasks such as semantic segmentation~\cite{hawke2020urban, jain2023ground}, object detection \cite{chitta2022transfuser, jaeger2023hidden}, and affordance prediction \cite{shao2023safety, li2018rethinking} provide meaningful intermediate representation for decision-making.
However, these methods do not engage end-users in the dialogue as they are one-directional and not readily comprehensible by the general users for the purpose of fostering trust and confidence.
An alternative promising approach is the integration of natural language explanations \cite{kim2018bddx, jin2023adapt, marcu2023lingoqa}, in particular through  Multi-Modal Large Language Models (MLLMs) \cite{achiam2023gpt, team2023gemini}.
These models, pre-trained on extensive web-scale datasets, demonstrate remarkable reasoning capacity, enabling the transformation of complex vehicular decision-making processes into more understandable narrative formats, thereby offering a new layer of explainability to conventional systems.


While several early attempts have demonstrated the potential of MLLMs as general explainable driving agents \cite{xu2023drivegpt4, wang2023drivemlm, ma2023dolphins}, these methods fall short of human-level understanding. One of the limitations is their failure to generalise to unseen environments. A primary obstacle is the lack of high-quality annotated data \cite{nie2023reason2drive}, coupled with the significant domain shift across various datasets \cite{domainbed}, which hinders \fixM{the model's}{2.7.14} generalisation capacity to novel environments outside of the training data distribution.
Another critical challenge is the prohibitively expensive training requirement and the unsolved problem of catastrophic forgetting \cite{kirkpatrick2017overcoming}, which make re-training or fine-tuning impractical solutions due to the immense computational demands and severe performance degradation. Consequently, this further limits the models' generalisability after deployment, as they struggle to effectively utilise new data in constantly evolving environments and driving scenarios.

To address these challenges, we introduce \textit{RAG-Driver}, a novel retrieval-augmented multi-modal large language model tailored for generalisable and explainable end-to-end driving.
As illustrated in \fixM{\cref{fig:front}}{2.7.16}, it outputs natural language texts corresponding to \textbf{(1)} the driving action and \textbf{(2)} justification of that driving action along with \textbf{(3)} numerical control signals based on driving videos. 
\fixM{Action and justification text accuracy, on the right, are much improved by our method (note that, for visualisation, in-distribution and out-of-distribution Performance have different vertical scales, with the latter being a much more difficult task).}{1.3.2}
The natural language texts are also aligned with control signals during in-context learning \cite{brown2020language} to enable faithful introspective explanation provision.
The novelty of \textit{RAG-Driver} is the integration of retrieval-augmented in-context learning (RA-ICL) mechanisms that significantly improve generalisation performance in unseen driving environments. It allows efficient recall of similar driving scenarios as contextual information augmenting MLLM prediction through implicit meta-optimisation (\cref{sec:RAGICL}). 
Through extensive experiments, we show that \textit{RAG-Driver} outperforms existing methods on both in-domain deployments as well as deployment in unseen environments (without any fine-tuning).
By being grounded in analogical demonstrations, our framework significantly reduces the need for continuous retraining while enhancing the generalisability and quality of generated explanatory texts.
Our primary contributions are as follows:
\begin{enumerate}
\item Proposing a novel retrieval-augmented in-context learning method for Multi-Modal Large Language Model (MLLM) based generalisable and explainable driving.
\item Achieving state-of-the-art introspective driving explanation performance on the standard benchmark BDD-X~\cite{kim2018bddx}.
\item Demonstrating exceptional zero-shot generalisation capacity to unseen scenarios without training effort through a customised dataset \textit{Spoken-SAX} featuring video sequences narrated by a professional driving instructor.
\end{enumerate}

\section{Related Work}

\subsection{Explainable End-to-End Autonomous Driving}

End-to-end learned driving~\cite{chen2023end} maps directly from raw sensor input to vehicle control signals.
This data-driven, joint optimisation of perception, prediction, and planning can be simple and efficient~\cite{Hu_2023_CVPR}. 
In this area, various learning-based approaches, including behavior cloning~\cite{bojarski2016end, chen2020learning, pomerleau1988alvinn, zhang2022learning, peng2023learning}, inverse optimal control~\cite{zeng2019end, sadat2020perceive, wang2021end}, and reinforcement learning~\cite{kendall2019learning, toromanoff2020end, chekroun2023gri, li2022efficient} are promising. 
A key area of focus in this area is explainability~\cite{zablocki2022explainability}, which is crucial for improving transparency and building trust towards wider public acceptance of autonomous systems~\cite{marcu2023lingoqa，gadd2020sense, OmeizaExplainSurvey2022}.
One line of works leverages attention visualisation -- either to directly identify salient regions of input images that are important for driving decision-making~\cite{kim2017interpretable, bojarski2016visualbackprop, bojarski2017explaining} or to assist feature aggregation for downstream motion planning tasks~\cite{moriVisualExpl, chitta2021neat, xiao2023scaling, renz2022plant}. 
Another line of work uses intermediate auxiliary tasks such as semantic segmentation~\cite{hawke2020urban, jain2023ground}, object detection \cite{chitta2022transfuser, jaeger2023hidden}, and affordance prediction \cite{shao2023safety, li2018rethinking} which help to \fixM{decode}{2.7.4} latent representations to human-understandable representation. 
While these methods provide explainable mechanisms by associating decision-making processes with semantic or visual representations, they are not readily comprehensible by the general users for the purpose of fostering trust and confidence.

Alternatively, recent research shows promise in utilising natural language explanation. 
Several works develop specialist explainers~\cite{kim2018bddx} using 
attention alignment in visual input and textual generation for grounded driving action explanation. ADAPT~\cite{jin2023adapt} uses a vision-language transformer with separate decoder for caption generation alongside control signal prediction. 
More recently, several works explore the potential of Multi-modal Large Language Models (MLLMs).
Works such as, DriveGPT4~\cite{xu2023drivegpt4},
Lingo~\cite{marcu2023lingoqa}, and DrivingMLM~\cite{wang2023drivemlm} have shown promising potential in general question-answering for driving and action planning.
However, a common obstacle in both specialist and MLLM-based generalist models is the scarcity of data due to expensive annotation costs and significant domain gaps between different datasets, making the development of a robust and generalisable model an extremely challenging task. 
In our work, we aim to overcome these obstacles by employing a more robust inference paradigm of retrieval-augmented in-context learning to bridge domain gaps and circumvent the need for annotations in new domains.

\begin{figure*}[t]
\centering
\includegraphics[width=\textwidth]{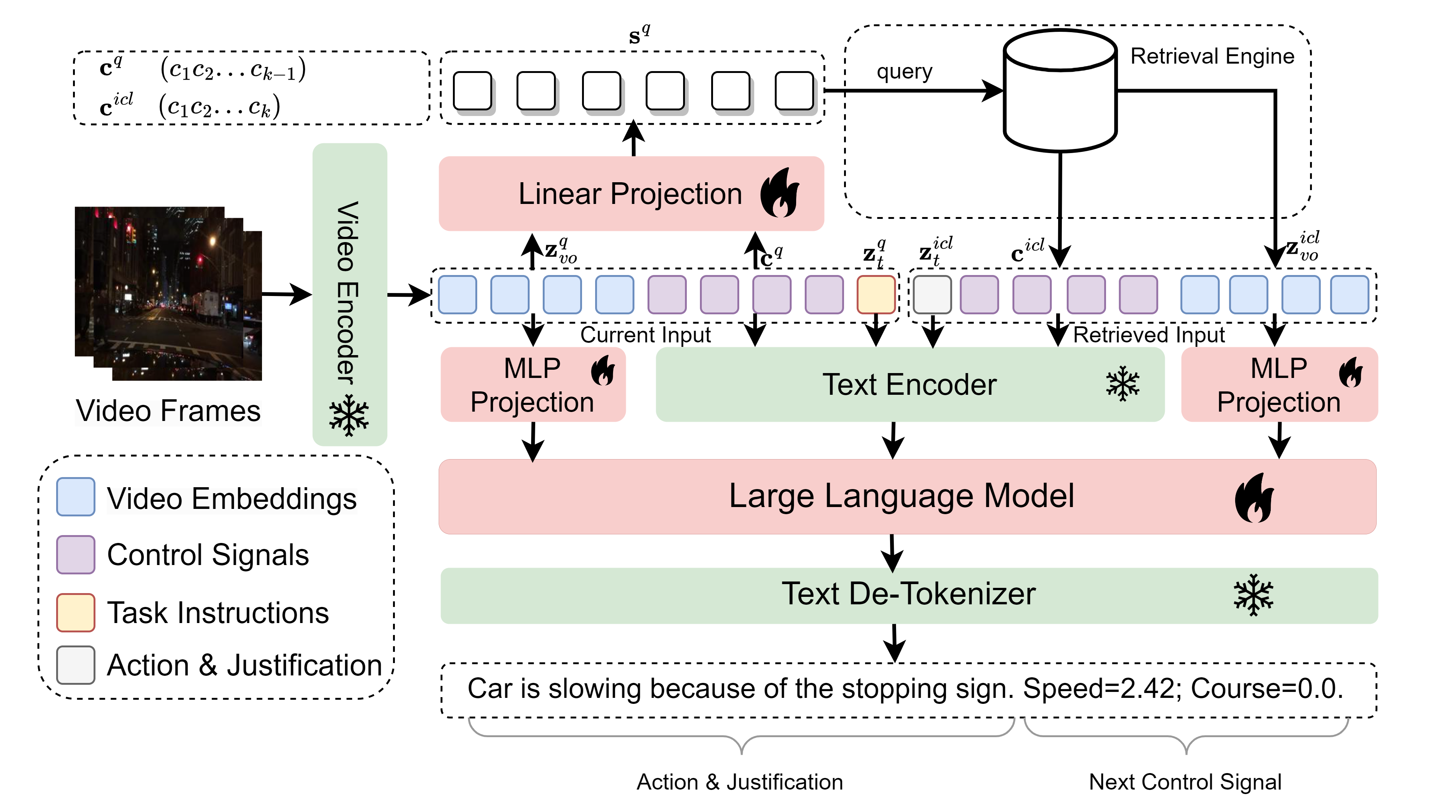}
\setcounter{figure}{1}
\caption{{\bf \textit{RAG-Driver} Overview:} Given a query comprising a video of the current driving scenario and its corresponding control signal, the process starts with the input being fed into a Retrieval Engine. It searches within a memory database for driving experiences that are similar to the current scenario, thereby providing relevant In-context Learning (ICL) samples. Subsequently, a Multi-Modal Large Language Model (MLLM) processes both the current query and the retrieved in-context learning samples, \fixM{where the encoded video embedding ($\bm{z}^q_{vo}$ or $\bm{z}^{icl}_{vo}$) is further projected through a MLP to align it with the language embedding ($\bm{z}^q_t$, $\bm{z}^{icl}_t$, or $\bm{c}^{icl}$) from the text encoder, so that the model can learn to associate relevant visual and textual features more effectively}{2.2}. Based on high-level task instructions, the model engages in various prediction tasks: Action Explanation, Action Justification, and Next Control Signal Prediction.}
\label{fig:main}
\vspace{-8mm}
\end{figure*}

\subsection{Multi-Modal Large Language Model}
Recent advancements in Large Language Models (LLMs) have paved the way for the emergence of Multi-modal Large Language Models (MLLMs)~\cite{achiam2023gpt, team2023gemini, alayrac2022flamingo, liu2023llava, zhu2023minigpt, li2023blip, he2024efficient, lin2023video, damonlpsg2023videollama}. Benefiting from scalable transformer-based architecture and web-scale training data, these models have demonstrated notable capabilities in general-purpose visual understanding tasks. 
One line of work focuses on modality fusion in the latent space, offering a scalable, end-to-end solution for MLLMs.
For instance, Flamingo~\cite{alayrac2022flamingo} and BLIP2~\cite{li2023blip} fuse visual tokens to frozen LLM through gated attention and query transformers, respectively. LLaVA~\cite{liu2023llava} and MiniGPT4~\cite{zhu2023minigpt} use simple Multi-Layer Perceptron (MLP) with visual instruction tuning to align the pre-trained image encoder to LLM. Most relevant to us is the line of work focusing on video-language models such as Video-LLaVA~\cite{lin2023video} and Video-LLaMA~\cite{damonlpsg2023videollama}, which integrate pre-trained video encoders into LLMs using strategies similar to those in image-based models.

With remarkable perception and reasoning capacity, 
MLLMs reveal promising potential in various robotics tasks such as reasoning~\cite{driess2023palm, ahn2022can, huang2023instruct2act} and planning~\cite{rana2023sayplan, brohan2023rt,lin2023text2motion}. 
The most similar to us is the idea that engages a generalist foundation model for end-to-end embodied agent. PaLM-e~\cite{driess2023palm} injects images, state estimates, and other sensor modalities to LLM and autoregressively produces natural language commands. RT-2~\cite{brohan2023rt} and RT-X~\cite{padalkar2023open} fine-tune on image and low-level robot control signal pairs to perform end-to-end robot control. 
Specifically in driving, numerous approaches leverage language-only \fixM{LLMs}{2.7.17} for decision-making \cite{yang2023survey} and then augment it with external perception module feedback~\cite{mao2023language,han2024dme, chen2023driving}, designed chain-of-thought reasoning \fixM{templates}{2.7.18}~\cite{mao2023language, ma2023lampilot} or downstream \fixM{planning}{2.7.19}~\cite{sha2023languagempc, sharan2023llm, chen2023driving} to form a system-level driving agent. Another more relevant line of work is end-to-end driving agents. DriveGPT4~\cite{xu2023drivegpt4} leverages a fine-tuned video-language model Valley~\cite{luo2023valley} on driving specific visual instruction tuning based on BDD-X~\cite{kim2018bddx}. Dolphins~\cite{ma2023dolphins} further use the designed Grounded chain-of-thought to enhance reasoning capacity. DrivingMLM~\cite{wang2023drivemlm} and Reason2Drive~\cite{nie2023reason2drive} scale up the driving visual instruction tuning dataset through simulator-based data engine and annotation of existing large-scale datasets, respectively. While these approaches have demonstrated the potentials of MLLM, the prohibitively expensive training cost and the unsolved problem of catastrophic forgetting, which makes re-training or fine-tuning post-deployment challenging, further \fixM{limiting}{2.7.20} their generalisation capacity to unseen driving environments. To solve this problem, we leverage a training-free retrieval-augmented in-context learning mechanism.

\subsection{In-Context Learning and Retrieval-Augmented Generation}

While LLMs demonstrate strong generative and reasoning capacity, there are still several issues associated with their output, such as hallucination~\cite{huang2023survey}, and slow knowledge updates~\cite{kandpal2023large}. In-context Learning (ICL)~\cite{brown2020language, dong2022survey} has emerged as a promising approach in LLM inference, potentially addressing several of these issues. This paradigm involves providing a test query alongside a few demonstration examples as contextual information. The LLM then generates an output for the test instance based on analogies drawn from context, without any updates to its parameters~\cite{rubin2021learning}. While ICL has been observed to enhance generalisability in various Natural Language Processing (NLP) tasks, its application in multi-modal contexts remains less explored, potentially due to the challenges associated with curating structured\fixM{,~}{2.7.21}high-quality multi-modal ICL datasets. 
Retrieval-Augmented Generation (RAG)~\cite{lewis2020retrieval} is another important inference paradigm for LLM. It provides an external knowledge database to augment the compressed \fixM{knowledge within the LLM}{2.7.22} in inference by dynamically retrieving relevant information \fixM{samples}{2.7.23} as contextual information. One of its promising applications is leveraging a systematic approach to curate In-Context Learning (ICL) examples.
In this work, we build upon these inference paradigms and extend their application to Multimodal Large Language Models (MLLMs). We introduce a retrieval-augmented in-context learning mechanism through 
a curated multi-modal driving in-context instruction tuning dataset and a vector similarity-based retrieval engine specifically tailored for driving applications.

\section{Method}

\textit{RAG-Driver} is a retrieval-augmented, multi-modal large language model (MLLM) for generalisable\fixM{,~}{2.7.24}explainable end-to-end driving. Its multi-tasking capabilities encompass three key areas: (1) Action Explanation, providing a human-understandable driving action description; (2) Action Justification, elucidating the rationale behind specific driving actions; and (3) Next Control Signal Prediction, forecasting upcoming control signals in response to the driving conditions. As shown in~\cref{fig:main}, it is composed of two primary components: a unified perception and planning unit built upon an MLLM backbone and a memory unit built upon a hybrid vector and textual database. These components interact through a retrieval engine, enabling robust multi-modal in-context learning (ICL) during decision-making. 

\subsection{Multi-modal Large Language Model Architecture}
\label{sec:architecture}
Following the successful MLLM paradigm of Video-LLaVA~\cite{lin2023video}, we align visual and language \fixM{embeddings}{2.7.25} through visual instruction tuning. We leverage a pre-trained video encoder and LLM and then inject the video embedding into \fixM{the}{2.7.26} LLM through \fixM{an MLP}{2.7.5.2} projector to build a fully differentiable MLLM.

\begin{figure}
\centering
\includegraphics[width=0.9\columnwidth]{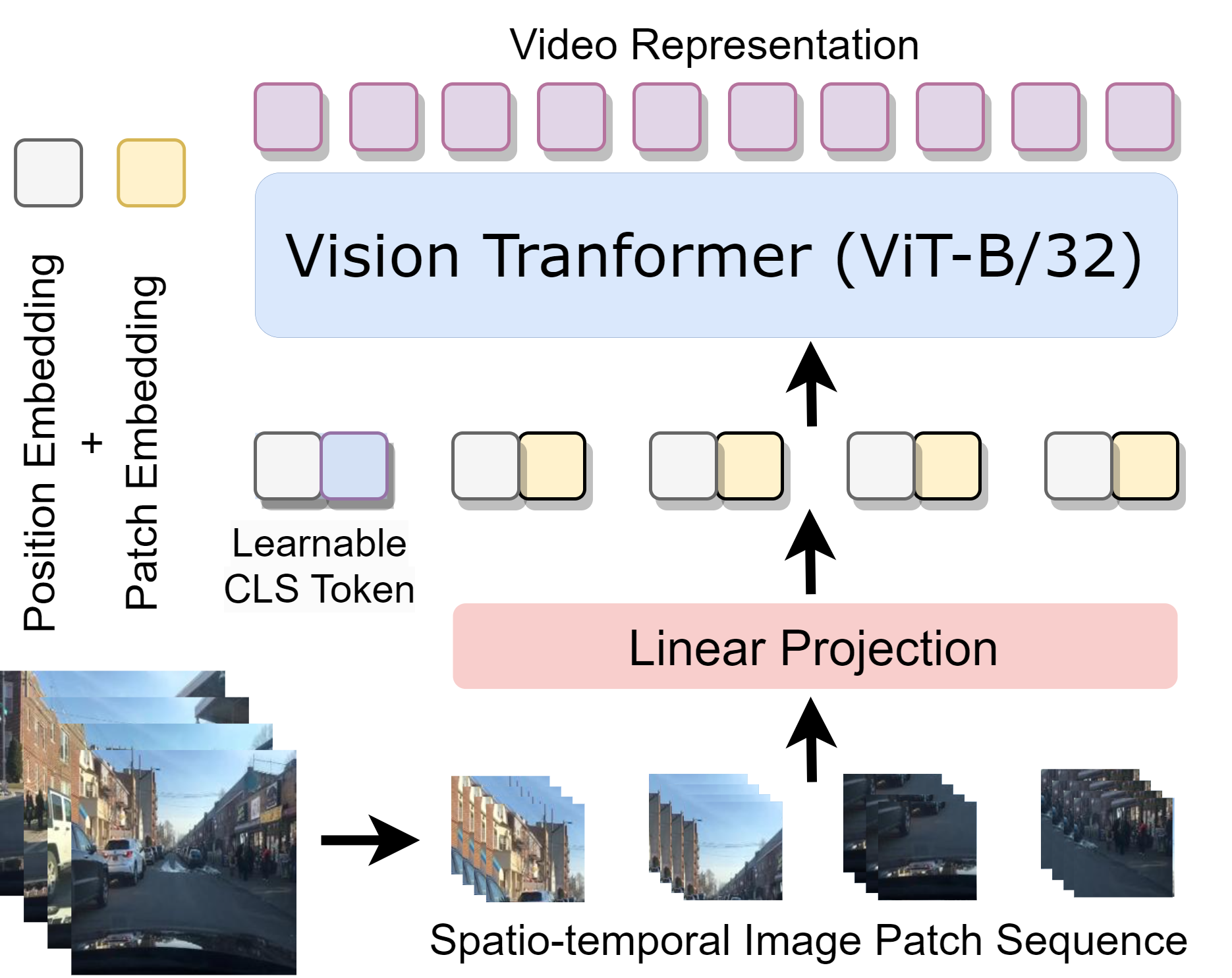}
\caption{{\bf Video Encoder architecture.} 
Video is first split into $k \times 32 \times 32$ patches concatenated in time, where these patches are linear projected to video embedding. Then, the model is trained with video-language contrastive learning (CLIP4clip)~\cite{Luo2021CLIP4Clip} to obtain \fixM{a language-aligned}{2.7.7} video representation.
}
\label{fig:vit}
\end{figure}

\noindent\textbf{Video Encoder} We adopt the pre-trained LanguageBind video encoder~\cite{zhu2023languagebind} as our frozen visual backbone \(f_v\), which is based on a ViT-B/32 vision transformer~\cite{dosovitskiy2020image}. As shown in \cref{fig:vit}, 
given an input video frame sequence \( V_{i} = \left\{ v_{i}^{1}, v_{i}^{2}, \ldots, v_{i}^{k} \right\} \in \mathbb{R}^{3 \times k \times 224 \times 224} \), we first split the video into multiple temporal sequences, each consisting of patches that share the same spatial location across different frames. These patches are then transformed through a linear projection for a vision transformer to output video embedding \( \bm{z}_{vo} \in \mathbb{R}^{2048 \times 1024} \). The video encoder is pre-trained with video-language contrastive learning (i.e. CLIP4clip~\cite{Luo2021CLIP4Clip}) without further fine-tuning.

\noindent\textbf{Cross-Modality Projector}
We then leverage a two-layer MLP to project and align the encoded video embedding \( \bm{z}_{vo} \) with language token embeddings \( \bm{z}_{v} \in \mathbb{R}^{ 2048 \times 4096}\). 
\begin{equation}
f_{p}(\bm{z}_{vo}) = \text{GELU}\left( W_2 \cdot \text{GELU}\left( W_1 \cdot \bm{z}_{vo} \right) \right)
\label{eq:projector}
\end{equation}
In particular, the projector \( f_{p}\) takes \fixM{the}{2.7.27} form in \cref{eq:projector}, where we use GELU~\cite{hendrycks2016gaussian} as an activation function. We train the projector with a two-stage training strategy as detailed in \cref{sec:training}.

\noindent\textbf{Large Language Model Backbone}
Finally, the LLM takes in both aligned video embedding \( \bm{z}_{v} \) and language embedding \( \bm{z}_{t} \) of textual context information and task instruction to predict \fixM{both the textual action explanation and the numerical control signal}{2.7.5.1}. We adopt Vicuna 1.5 7B~\cite{zheng2023judging}, which is instruction-tuned based on LLaMA2~\cite{touvron2023llama2} as our LLM backbone. 
For \fixM{the}{2.7.28} decoder-only LLM conditioned on the multi-modal contextual prefix \( \bm{z}_{1:n} = [\bm{z}_{v},\bm{z}_{t}] \) of length \(N\), the joint probability of \fixM{the output \(\bm{x}_{n+1:L}\) is as per}{2.7.29} \cref{eq:llm}, where \( P_{\theta} \) is \fixM{the}{2.7.30} transformer-based LLM backbone parameterized by \( \theta \). 
\begin{equation}
P(\bm{x}_{n+1:L} \mid \bm{z}_{1:n}) = \prod_{l=n+1}^L P_{\theta}(\bm{x}_l \mid \bm{x}_{1:l-1}, \bm{z}_{1:n})
\label{eq:llm}
\end{equation}
Each output token \(\bm{x}_{l}\) is then sampled auto-regressively based on \fixM{the}{2.7.31} previous output and context and finally decode to language space through a text de-tokenizer.

\subsection{Training Strategy}
\label{sec:training} 
Following the visual instruction tuning paradigm~\cite{liu2023llava, lin2023video}, we employ a two-stage training strategy to progressively enable cross-modality alignment and multi-task driving capacity. In both stages, we leverage the same next-token prediction cross-entropy loss as in \cref{eq:loss} aiming to maximize the conditional answer likelihood in \cref{eq:llm}, where \( \bm{y} \) is the ground truth token.
\begin{equation}
\mathcal{L}_{CE} = -\sum_{i=n+1}^{L} \bm{y}_{l} \log P(\bm{x}_{l} \mid \bm{z}_{1:n})
\label{eq:loss}
\end{equation}

\noindent\textbf{Pre-training}
In the first pre-training stage, we only train the cross-modality projector while freezing the visual encoder and LLM. A subset \fixM{of}{2.7.32} VIDAL-10M~\cite{zhu2023languagebind} which contains 3 million video-caption pairs. This achieves \fixS{an}{2.7.33} alignment between visual and language features by projecting the pre-trained video embeddings into language tokens understandable by LLM.

\begin{figure*}
\centering
\includegraphics[width=\textwidth]{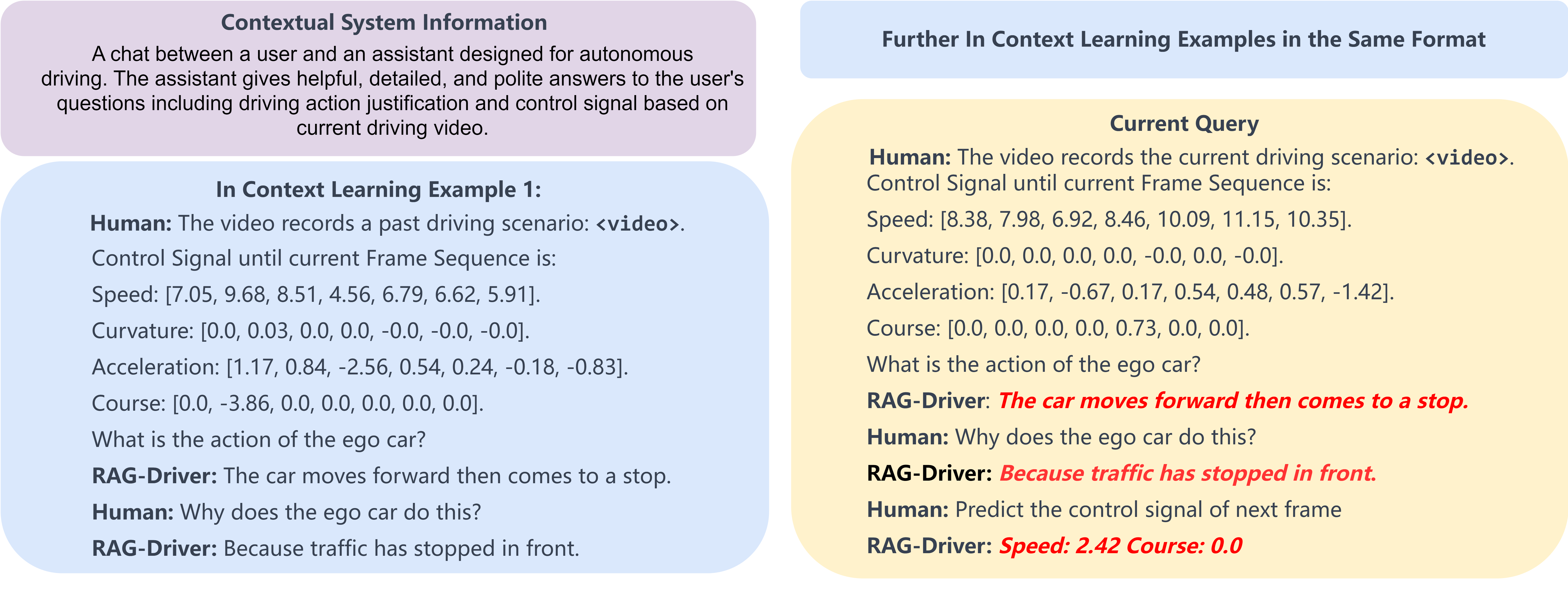}
\caption{{\bf Query Sample}. A single query includes three primary components: (1) Contextual System Information, a constant element across all queries that offers high-level task instruction; (2) Current Query, comprising all driving information and specific task instruction in natural language question-answer format. \fixM{Driving video visual inputs (i.e. \textbf{$<$video$>$}) as represented here are in fact encoded video tokens from a video encoder and MLP projector which are concatenated with language tokens from a text encoder, see~\cref{fig:main} for more architectural detail.}{2.4} Note that the red sentence is the expected output from MLLM; (3) Multiple In-Context Learning Examples, retrieved from memory, which serve as an analogical demonstration of a complete reasoning process—from sensor input in driving scenarios to action justification and control prediction.}
\label{fig:sample}
\vspace{-6mm}
\end{figure*}

\noindent\textbf{Instruction Tuning}
While state-of-the-art \fixM{LLMs exhibit}{2.7.52} zero-shot ICL capability, several works~\cite{dong2022survey} and our ablation study \fixM{in}{2.7.45} \cref{sec:ablation} have shown further improvement when specifically trained on curated ICL demonstrations. In particular, we construct a multi-modal instruction tuning dataset with structured ICL examples based on the BDD-X dataset~\cite{kim2018bddx} resulting in 16K video question and answer pairs. As shown in \cref{fig:sample}, for an 8-frame driving video sequence with associated control signals \fixM{--}{2.7.42} speed, course, acceleration, and curvature as the current query, \fixS{we use the retrieval mechanism as in~Sec. III-C}{2.3.1}\fixN{ we retrieve two relevant driving experiences}{2.3.2}, which are then prefixed to the current query as ICL examples. The dataset is tailored to support three distinct tasks, each represented through question-answer pairs in natural language. Note that, with (1) Action Explanation and (2) Justification naturally represented as natural language, (3) Control Signal prediction is also formed as language token prediction; this is feasible due to the distinct mapping of numerical values to specific tokens within the language model dictionary. 

\subsection{Retrival-Augmented In-context Learning}
\label{sec:RAGICL}
Another critical component of the system is the memory unit, which consists of a database and a retrieval engine. The database \fixM{takes in a}{2.7.34} vectorised video embedding \( \bm{z}_{vo} \), which is extracted with the same video encoder as in \cref{sec:architecture}, and control signals \( \bm{c} \in \mathbb{R}^{28}\) directly from sensor recording. Each vector is uniquely associated with the corresponding human expert textual explanation and justification from \fixM{training}{2.7.35} samples as in \cref{sec:training}. 

\noindent\textbf{Retrieval Mechanism}
To perform the retrieval, we first leverage a lightweight MLP projector of the same structure as in \cref{eq:projector} to project the heterogeneous video and control signal embedding into the same hybrid embedding $\bm{s} \in \mathbb{R}^{1024}$ through metric learning \cite{hoffer2015deep}. In particular, we adopt a triplet loss with Euclidean distance as shown in \cref{eq:tripletmargin}:
\begin{equation}
\mathcal{L}_{Tri}(\bm{a}, \bm{p}, \bm{n}) = \operatorname*{max}(||\bm{a} - \bm{p}||_2 - ||\bm{a} - \bm{n}||_2 + \operatorname{margin}, 0)
\label{eq:tripletmargin}
\end{equation}
\fixM{where $\bm{a}$, $\bm{p}$, $\bm{n}$ defines anchor, positive, and negative data sample, respectively.}{3.7} The positive $(\bm{a},\bm{p})$ and negative pairs $(\bm{n},\bm{p})$ between hybrid embedding $\bm{s}$ are selected \fixM{}{2.7.36} based on text similarity (i.e., TF-IDF Score) \fixM{of the}{} driving action and justification in the BDD-X training set\fixM{. We}{2.7.8} aim to form the metric space such that the scenarios lead to similar driving actions close together and vice versa. 
This approach addresses the limitations of relying solely on visual similarity, which we have empirically found can result in sub-optimal performance (\cref{sec:ablation}), and also solves the problem of heterogeneous sensor inputs that are \fixM{difficult}{2.7.37} for similarity comparison. We then perform retrieval through an efficient vector similarity search.
Given a query vector $\bm{s}^{q}$\fixM{, }{2.7.38} the cosine similarity between the query vector and each vector in the database is computed as 
\fixM{$S_{c}(\bm{s}_{q}, \bm{s}^{(i)}) = \frac{\bm{s}_{q} \cdot \bm{s}^{(i)}}{\|\bm{s}_{q}\| \|\bm{s}^{(i)}\|}$.}{2.7.6}
Subsequently, we consistently select the two most relevant driving samples based on this similarity score. These samples represent the entire reasoning process, from contextual information to question-answer pairs, as illustrated in \cref{fig:sample}. 

\noindent\textbf{Retrieval Augmented In-Context Learning (RA-ICL)}
To perform the RA-ICL, we prefix \fixM{these}{2.7.39} retrieved samples before the current query, facilitating an implicit gradient descent through the meta-optimiser of LLM, as proved in~\cite{dai2022can}. This approach is also applicable to MLLM\fixM{s, with our MLLM}{2.7.40} architecture specified in \cref{sec:architecture}.
For a given transformer-based pre-trained MLLM modeling the prefix-conditioned output probability as in \cref{eq:llm}, consider one head of the multi-head attention in a single transformer block as follows:
\begin{equation}
\begin{aligned}
\mathcal{F}_{\mathrm{ICL}}&(\bm{z}_{1:n})  =
\operatorname{Attention}(Q, K, V) \\
= & W_V\left[\bm{z}_{icl} ; \bm{z}_{q}\right] \operatorname{Softmax}\left(\frac{\left(W_K\bm{z}_{1:n}\right)^T (W_Q\bm{z}_{1:n})}{\sqrt{d^{i}}}\right)
\end{aligned}
\label{eq:icl_1}
\end{equation}
where \( \bm{z}_{1:n} = \left[\bm{z}_{icl} ; \bm{z}_{q}\right]\) \fixM{represents}{2.7.41} the ICL example $\bm{z}_{icl}$ and current query embeddings $\bm{z}_{q}$, respectively. $W_{Q,K,V} \in \mathbb{R}^{d^{i} \times d^{o}}$ is the linear transformation on query, key, and value \fixM{in the}{2.7.43} attention block, respectively.
Now, in~\cite{dai2022can} in-context learning was shown to effectively achieve a meta-optimisation to implicitly update the model with an estimated meta-gradient.
We provide in our work a new, alternative derivation of this, supported by more recent work in~\cite{koohpayegani2024sima} on the following \textit{softmax-free} linear attention expressions
\begin{equation}
\begin{aligned}
\mathcal{F}_{\mathrm{ICL}}(\bm{z}_{1:n}) = W_V\left[\bm{z}_{icl} ; \bm{z}_{q}\right]\left(W_K\left[\bm{z}_{icl} ; \bm{z}_{q}\right]\right)^T W_Q\bm{z}_{1:n}
\end{aligned}
\end{equation}
This can be simplified, with a more detailed derivation in~\cref{apx:ICL_Derivation}, as
\begin{equation}
\begin{aligned}
\mathcal{F}_{\mathrm{ICL}}(\bm{z}_{1:n})=
(\Delta{}W_{ICL}+W_{ZSL})W_Q\bm{z}_{1:n}
\end{aligned}
\end{equation}
where we separate out the terms $W_{ZSL}$ independent of the ICL examples and dependent solely on the current query from those $\Delta{}W_{ICL}$ dependent on the ICL examples, given by:
\begin{equation}
\begin{aligned}
\Delta{}W_{ICL}&=\sum_i~W_V\bm{z}_{icl,i}(W_K\bm{z}_{icl,i})^T\\
W_{ZSL}&=W_V\bm{z}_{q}(W_K\bm{z}_{q})^T
\end{aligned}
\end{equation}
Now, with more detail in~\cref{apx:ICL_Derivation}, a forward-pass
\begin{equation}
\mathcal{F}(\mathbf{x})=(W_0+\Delta{}W)\mathbf{x}
\end{equation}through a linear layer $\mathcal{F}$ after its weights $W_0$ have been updated by $\Delta{}W$ has the weight updates coupled to the input in the form
\begin{equation}
\begin{aligned}
\Delta{}W\mathbf{x}= 
\sum_i\eta\frac{\partial{}L}{\partial\mathbf{y}}|_{\mathbf{y}_i}\mathbf{x}_i^T\mathbf{x}
\end{aligned}
\end{equation}
where $\mathbf{x}_i,\mathbf{y}_i$ are the (mini-batch) input and output to the layer that resulted in the weight update when backpropagating loss $L$ and with learning rate $\eta$.
Therefore we have a weighted sum of dot products, which is \textit{akin to an attention mechanism}.
Indeed, by inspection of similar dot-product expressions $W_V\bm{z}_{icl,i}(W_K\bm{z}_{icl,i})^T\leftrightarrow\eta\frac{\partial{}L}{\partial\mathbf{y}}|_{\mathbf{y}_i}\mathbf{x}_i^T\mathbf{x}$ we note that we match the form for the linear layer above
$$
(W_{ZSL}+\Delta{}W_{ICL})W_Q\bm{z}_{1:n}
\leftrightarrow
(W+\Delta{}W)\mathbf{x}
$$
This can therefore be interpreted to say that the output of the attention is adjusted in a meta-optimal way to conform to the samples provided as input context, much like gradient descent on the linear layer would adjust that layer to conform to the mini-batch training data, but \textit{crucially} in the case of \textit{RAG-Driver}, without backpropagation.

RA-ICL serves as an efficient inference method boosting the performance of \fixM{MLLMs}{2.7.44} in explainable driving without further training effort, where we empirically verify it is extremely effective in boosting the model prediction performance and generalisation capacity.

\section{Experiments}
\label{sec:experiments}

\subsection{Settings and Datasets}
\label{sec:settings_datasets}
We empirically evaluate the proposed \fixM{framework of Retrieval-augmented In-Context Learning (RA-ICL) framework for Multi-modal Large Language Model (MLLMs)}{2.7.53}, targeting explainable driving applications. We aim to validate its efficacy in general driving scenarios with a focus on two main aspects:
\textbf{(1)} explainability in driving action explanation and justification. 
\textbf{(2)} Control Signal Prediction. We conduct experiments with the BDD-X~\cite{kim2018bddx} dataset, which is a widely adopted benchmark in explainable driving, comprising 77-hour videos across the US under different road and weather conditions. We customize the format as shown in~\cref{fig:sample}, resulting in 16,803 and 2,123 video question-answering pairs for training and testing, respectively. More importantly, we further explore the transfer learning capacity of
zero-shot generalisation in unseen environments. We leverage customised dataset
comprising 58 testing question-answering pairs, recorded in London, UK as part of the Sense-Assess-eXplain dataset~\cite{gadd2020sense}, presenting a significant distribution shift from the BDD-X dataset.

\noindent\textbf{Benchmark Settings} For all experiments, we train the MLLM using the BDD-X training split. Subsequent evaluation on general explainability and control signal prediction capabilities tests are conducted on the BDD-X test split with the BDD-X training split as the memory database. For the transfer learning experiments, we employ the same foundational model and test it on the \textit{Spoken-SAX} \fixM{dataset}{2.7.10.1} but the memory database is constructed using the BDD-X training split for zero-shot generalisation.

\noindent\textbf{Implementation \fixM{Details}{2.7.10.2}} for each of the driving videos, we uniformly sample the video to 8 frames and resize it to $224 \times 224$ for all frames. For MLLM, we train the model for one and two epochs in the pre-training and fine-tuning stages, respectively. For the embedding projector, we train the model for 300 epochs. 
\fixM{Training of the retrieval engine on the BDD-X dataset using a single A100 GPU takes half an hour. Then, we utilize the pre-training checkpoint from Video-LLaVA \cite{lin2023video} and fine-tune the MLLM on top of it, which takes 6 hours on 8 A100 GPUs. For the inference time, since we store all driving experience in a preprocessed database, which speeds up the retrieval process, the overall single round inference time is roughly 4 seconds on a single A100 GPU.}{3.5}
Further experiment implementation details are provided in \cref{apx:training_details}.

\noindent\textbf{Evaluation Metric} 
For the driving action description and justification tasks, we use the same metrics as~\cite{jin2023adapt} including 4-gram BLEU (B4)~\cite{papineni2002bleu}, METEOR (M)~\cite{banerjee2005meteor}, and CIDEr (C)~\cite{vedantam2015cider}. These metrics aim to evaluate text generation quality, with BLEU focusing on n-gram precision, METEOR incorporating semantic and syntactic nuances, and CIDEr \fixM{emphasising}{2.7.46} consensus and relevance in tasks like image captioning. Moreover, \fixM{for}{2.7.10.3} the control signal evaluation, we again follow~\cite{jin2023adapt} and present Root Mean Square Error (RMSE) in both steering angle (\SI{}{\degree}) and speed (\SI{}{\metre\per\second}).
We also present ``tolerant accuracy'' metrics, $A_\sigma$, representing the accuracy of predictions when binarised as within a tolerance threshold $\sigma$ of the ground truth.

\begin{table*}
\centering
\caption{
\textbf{Driving Action Explanation and Justification Evaluation.} B4, C, and M represent BLEU4, CIDEr, and METEOR, respectively, for measuring the accuracy and quality of textual driving explanation. The upper part indicates the evaluation on the in-distribution BDD-X test set. The lower part indicates the zero-shot generalisation evaluation on the Spoken-SAX dataset. 
}
\resizebox{.95\linewidth}{!}{%
\begin{tabular}{lcccccccc}
\hline 
\multirow{2}{*}{Method} & \multirow{2}{*}{Is Generalist?} & \multicolumn{3}{c}{Action} & & \multicolumn{3}{c}{Justification} \\
\cline{3-5} \cline{7-9}
& & B4 $\uparrow$ & C $\uparrow$ & M $\uparrow$ & & B4 $\uparrow$ & C $\uparrow$ & M $\uparrow$ \\
\hline 
\textit{In Distribution Performance (BDD-X)} \\
S2VT \cite{venugopalan2015sequence} & $\times$ & $30.2$ & $179.8$ & $27.5$ & & $6.3$ & $53.4$ & $11.2$ \\
S2VT++~\cite{venugopalan2015sequence} & $\times$ & $27.1$ & $157.0$ & $26.4$ & & $5.8$ & $52.7$ & $10.9$ \\
SAA \cite{kim2018bddx} & $\times$ & $31.8$ & $214.8$ & $29.1$ & & 7.1 & $66.1$ & $12.2$ \\
WAA~\cite{kim2018bddx} & $\times$ & $32.3$ & $215.8$ & $29.2$ & & $7.3$ & $69.5$ & $12.2$ \\
ADAPT~\cite{jin2023adapt} & $\times$ & $34.6$ & $247.5$ & $30.6$ & & $11.4$ & $102.6$ & $15.2$ \\
DriveGPT4~\cite{xu2023drivegpt4} & \checkmark & $30.0$ & $214.0$ & $29.8$ & & $9.4$ & $102.7$ & $14.6$ \\
\rowcolor{Gray!16}
OURS & \checkmark & $34.3$ & $260.8$ & $30.7$ & & $11.1$ & $109.1$ & $14.8$ \\ 
\rowcolor{Gray!16}
$\Delta$ \textit{w.r.t generalist SOTA} &  & \gain{$14.3\%$} & \gain{$21.9\%$} & \gain{$3.0\%$} & & \gain{$18.1\%$} & \gain{$6.2\%$} & \gain{$1.4\%$} \\
\midrule
\textit{Zero-shot Generalisation (Spoken-SAX)} & & B4 $\uparrow$ & C $\uparrow$ & M $\uparrow$ & & B3 $\uparrow$ & C $\uparrow$ & M $\uparrow$ \\
ADAPT~\cite{jin2023adapt} & $\times$ & $0.0$ & $22.3$ & $26.7$ & & $3.1$ & $8.6$ & $12.4$\\
Base w/o ICL & \checkmark & $0.0$ & $5.7$ & $21.6$ & & $2.1$ & $4.7$ & $11.4$\\
\rowcolor{Gray!16}
OURS w ICL & \checkmark & $9.9$ & $48.9$ & $27.5$ & & $5.6$ & $17.1$ & $14.6$\\
\rowcolor{Gray!16}
$\Delta$ \textit{w.r.t SOTA} &  & \gain{$-\%$} & \gain{$119.3\%$} & \gain{$3.0\%$} & & \gain{80.6\%} & \gain{$98.8\%$} & \gain{$17.7\%$} \\
\bottomrule
\end{tabular}}
\label{tab:benchmark_main}
\end{table*}

\begin{table*}
\begin{center}
\caption{\textbf{Control Signals Prediction Accuracy Evaluation on BDD-X dataset.}}
\label{tab:benchmark_signal}
\resizebox{\linewidth}{!}{%
\begin{tabular}{@{}ccccccccccccc@{}}
\toprule
\multirow{2}{*}{Method} & \multicolumn{6}{c}{Course} & \multicolumn{6}{c}{Speed} \\ 
\cmidrule(l){2-13} 
& RMSE(degree)$\downarrow$ & $A_{0.1}\uparrow$ & $A_{0.5}\uparrow$ & $A_{1.0}\uparrow$ & $A_{5.0}\uparrow$ & $A_{10.0}\uparrow$ & RMSE(m/s)$\downarrow$ & $A_{0.1}\uparrow$ & $A_{0.5}\uparrow$ & $A_{1.0}\uparrow$ & $A_{5.0}\uparrow$ & $A_{10.0}\uparrow$ \\
\midrule
ADAPT~\cite{jin2023adapt} & $5.87$ & $54.49$ & $86.39$ & $91.06$ & $97.36$  & $98.20$ & $2.68$ & $11.77$ & $31.79$ & $47.48$ & $92.75$ & $95.87$ \\
DriveGPT4~\cite{xu2023drivegpt4} & $4.57$ & $69.22$ & $79.14$ & $84.47$ & $95.72$ & $96.74$ & $1.09$ & $\textbf{56.93}$ & $77.77$ & $87.97$ & $99.00$ & $99.57$  \\
\midrule
\rowcolor{Gray!16}
OURS & $\textbf{4.48}$ & $\textbf{74.32}$ & $\textbf{88.69}$ & $\textbf{93.12}$ & $\textbf{98.30}$ & $\textbf{99.10}$ & $\textbf{0.69}$ & $51.12$ & $\textbf{85.54}$ & $\textbf{94.49}$ & $\textbf{99.81}$ & $\textbf{99.91}$ \\
\bottomrule
\end{tabular}
}
\end{center} 
\end{table*}

\begin{table*}
\centering
\small
\caption{{\bf Ablation Study} on Retrieval Strategy and In-Context Learning Phase. Visual and Hybrid Search refers to the use of video embeddings or projected hybrid sensor embeddings, respectively. Training and Inference denote whether in-context learning examples are provided during the corresponding phase. }
\label{tab:ablation}
\resizebox{0.8\textwidth}{!}{
\begin{tabular}{cc|cc|cccccc}
\toprule
\multicolumn{2}{c}{Retrieval Strategy} & \multicolumn{2}{c}{ICL Phase} & \multicolumn{6}{c}{BDD-X} \\
\midrule
Visual Search & Hybrid Search & Training & Inference & Act. B4 & Act. C & Jus. B4 & Jus. C & Speed Err. & Course Err.  \\ 
\midrule
&              &         &       &   $29.8$           &  $222.1$ & $5.9$ & $58.9$ & $0.67$ & $5.61$   \\
\midrule
\Checkmark            &              &              &  \Checkmark      &  $0.0$     & $0.0$  &$0.0$ & $0.0$ & $N/A$ & $N/A$   \\
             & \Checkmark            &              &   \Checkmark    &     $0.0$        & $0.0$  & $0.0$& $0.0$ & $N/A$ & $N/A$    \\
\Checkmark            &              &   \Checkmark          &  \Checkmark  &   $31.2$           &  $222.1$ & $7.7$ & $83.1$ & $0.78$ & $5.61$   \\

             & \Checkmark            &   \Checkmark          &  \Checkmark   &  $34.3$   & $260.9$ & $11.1$ & $109.1$ & $0.69$ & $4.48$ \\
 
\bottomrule
\end{tabular}
}
\end{table*}

\begin{figure*}[t]
\centering
\includegraphics[width=\textwidth]{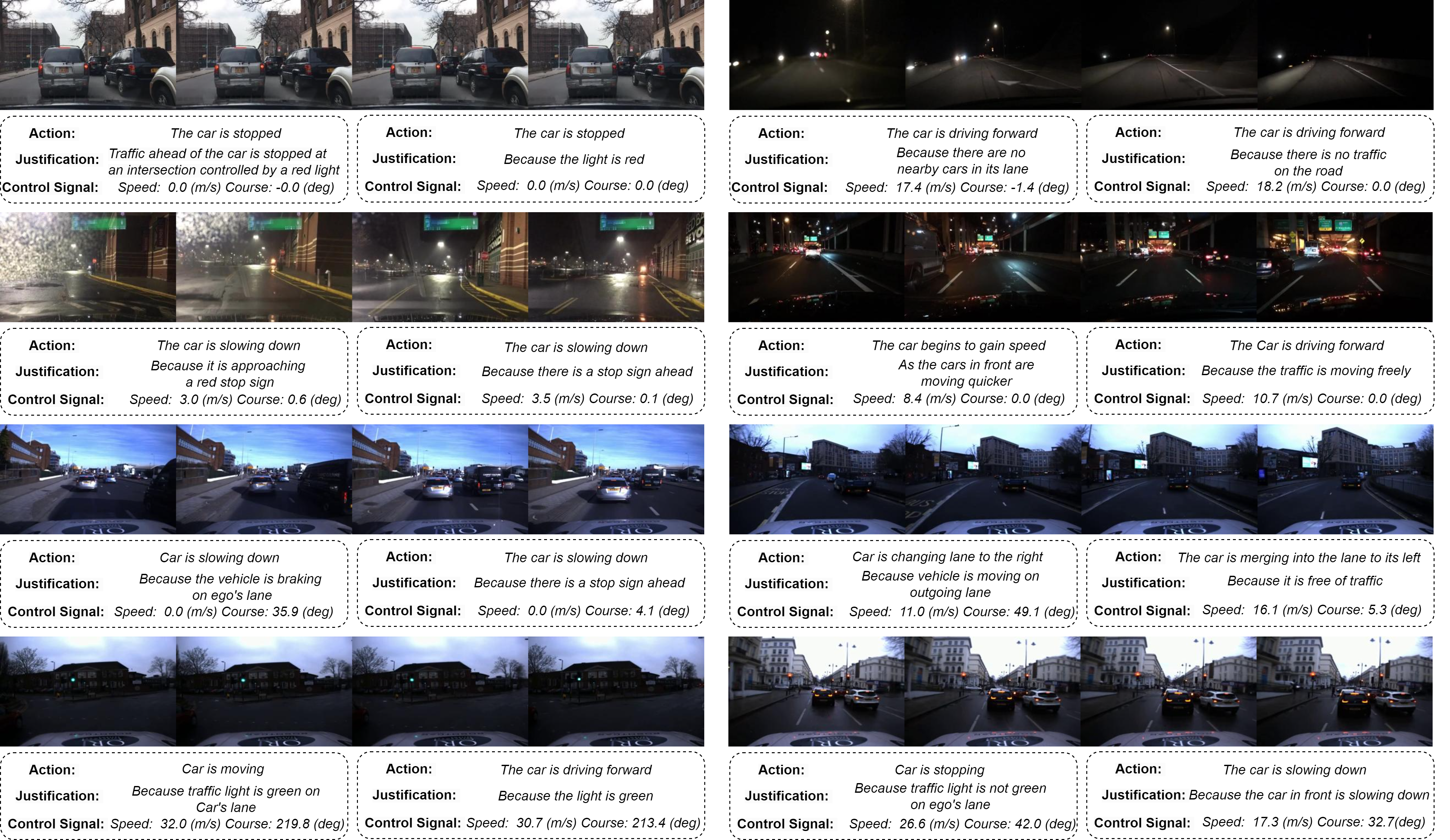}
\caption{{\bf Qualitative Examples.} Comparison between human annotated ground truth (\textbf{Left}) and prediction from our method (\textbf{Right}) from both BDD-X (\textbf{Top 4}) and \fixM{Spoken-SAX}{2.7.11} (\textbf{Bottom 4}). Each video is downsampled to four frames for visualisation. Despite subtle differences in wording, our method provides driving action explanations and justifications that are comprehensible to humans.}
\label{fig:demo}
\end{figure*}

\noindent\textbf{Baselines}
We compare with various driving action description and justification baseline methods such as video-language sequence-to-sequence based recurrent neural network S2VT~\cite{venugopalan2015sequence} and visual-attention based convolutional neural network WAA~\cite{kim2018bddx}. \fixM{We also compare with a state-of-the-art video transformer-based driving action explanation method ADAPT~\cite{jin2023adapt}.
Methods like this are specialist in that they use two branches for explanation and control signal prediction.
Beyond these specialist models for driving explanation, we further compare with the generalist visual instruction-tuned MLLM, DriveGPT4~\cite{xu2023drivegpt4}.
Generalist methods like this use the same backbone for explanation and control signal prediction.
}{2.6}

\subsection{Explainability in Driving Action and Justification}
We begin by evaluating the quality and accuracy of explanations and justifications for driving actions separately. As shown in upper part of \cref{tab:benchmark_main}, our method demonstrates comparable performance to the state-of-the-art specialist method ADAPT~\cite{jin2023adapt}, a characteristic not observed in previous MLLM-based methods. In particular, when compared with DriveGPT4~\cite{xu2023drivegpt4} which also uses \fixM{an}{2.7.47} MLLM with a similar architecture and number of parameters but incorporates the extra LLaVA-150K dataset~\cite{liu2023llava} for visual instruction tuning, our approach, relying solely on the BDD-X dataset, outperforms it in terms of explainability. This is evidenced by an average performance improvement of 10.8\% across all metrics. This underscores the effectiveness of ICL in enhancing the emergent reasoning capabilities of MLLMs.

\subsection{Control Signal Prediction}
\label{sec:control_signal}

We next evaluate the accuracy of control signal predictions for Course (i.e., turning angle) and Speed. As indicated in \cref{tab:benchmark_signal}, our method surpasses others in open-loop control accuracy across various tolerance ranges and in terms of RMSE, significantly outperforming the baseline approach. In particular, when compared to the state-of-the-art DriveGPT4, which also uses the same visual input combined with past control signals for autoregressive prediction, our method stands out by implementing retrieval-augmented ICL examples. This indicates \fixM{that}{2.7.48} the analogy in the overall reasoning process provided by the ICL examples also contributes to the improvement in numerical control signal prediction.

\subsection{Ablation Study on Retrieval Strategy}
\label{sec:ablation}

We perform a more comprehensive ablation study to evaluate the efficacy of our proposed retrieval-augmented in-context learning.
We first aim to investigate the similarity metric for retrieval. In particular, we compare the use of visual similarity (i.e., video embeddings only) with hybrid similarity (i.e., hybrid video and control signal projected embedding \cref{sec:RAGICL}). Our empirical findings indicate suboptimal performance when using visual similarity, possibly because it tends to \fixM{prioritise}{2.7.49} ICL examples that are most perceptually similar, rather than effectively demonstrating the reasoning process. By fine-tuning the embeddings, we not only harness the potential of heterogeneous multi-modal sensor input but also enable more effective ICL example retrieval.

We also investigate whether to apply ICL examples during training or solely in inference time.
As shown in \cref{tab:ablation}, we found that the MLLM is incapable of making reasonable predictions using ICL examples without prior training, regardless of the retrieval strategy chosen; \fixM{as the model generates random string and non-floating-point number, which precludes the calculation of sensible metrics indicated as N.A in the corresponding entry.}{3.3} This suggests that the pre-trained MLLM is not equipped to effectively perform zero-shot ICL. We hypothesise that supervised fine-tuning plays a crucial role in enhancing the ICL capabilities of MLLM, necessitating the provision of reasoning demonstrations, which aligns with observations in~\cite{dong2022survey}.

\fixM{
\begin{table}
\centering
\small
\caption{{\bf Ablation Study} on number of in-context learning examples on BDD-X.}
\label{tab:ablation_nicl}
\resizebox{0.95\columnwidth}{!}{
\begin{tabular}{c|cccccc}
\toprule
 No. ICL Examples & Act. B4 & Act. C & Jus. B4 & Jus. C & Speed Err. & Course Err.  \\ 
\midrule
1 & $33.2$ & $257.2$ & $10.2$ & $99.3$ & $0.75$ & $4.15$ \\
2 & $34.3$ & $260.9$ & $11.1$ & $109.1$ & $0.69$ & $4.48$ \\
\bottomrule
\end{tabular}
}
\end{table}

We further investigate the effect of the number of ICL examples. In particular, we compare the performance using the same optimal hybrid search as above but with a varying number of ICL examples. As shown in \cref{tab:ablation_nicl}, we observe that increasing the number of ICL examples from $1$ to $2$ improves performance in explanation and justification while slightly decreasing the course control signal accuracy.
It is worth noting that the choice of at most two ICL examples is constrained by the MLLM architecture. We include analysis and discussion on this limitation in \cref{sec:lim_scale}.}{1.4.1}

\subsection{Generalisation Capacity\label{sec:generalisation}}
One of the critical capacities of autonomous systems is to generalise to unseen environments out of its training distribution. However, in the domain of explainable driving, we observe the existing methods are unable to perform such generalisation, which poses challenges for their deployment. As shown in lower part of \cref{tab:benchmark_main}, ADAPT and the base MLLM (i.e., training without ICL) reveal dramatic performance degradation \fixM{as compared}{2.7.50} to the in-distribution situation. However, our method leverages ICL learning examples to demonstrate a significant performance boost with a large margin. Note that even though the memory database is constructed with BDD-X, the RA-ICL can still perform generalisation in a zero-shot manner. This is potentially due to the robustness of the hybrid retrieval process, where samples with less distribution shift can still be selected to serve as effective ICL examples.

\fixM{\subsection{Fine-tuning Retrieval Engine\label{sec:fine-tuning-experiments}} 
\begin{table*}
\centering
\small
\caption{Effect of training retrieval engine with target domain data samples.}
\label{tab:engine_finetune}
\begin{tabular}{c|cccccccc}
\toprule
 & \multicolumn{4}{c}{BDD-X} & \multicolumn{4}{c}{Spoken-SAX} \\
\midrule
 Finetune? & Act. B4 & Act. C & Jus. B4 & Jus. C  & Act. B4 & Act. C & Jus. B3 & Jus. C  \\ 
\midrule
$\times$ & $34.3$ & $260.9$ & $11.1$ & $109.1$ & $9.9$ & $48.9$ & $5.6$ & $17.1$ \\
\checkmark & $34.7$ & $271.0$ & $10.9$ & $112.4$ & $10.1$ & $52.1$ & $6.7$ & $19.0$  \\
\bottomrule
\end{tabular}
\end{table*}

While RAG-Driver aims to enhance the model's generalisability in driving action explanation tasks in a training-free manner through in-context learning, we further investigate the potential to improve the retrieval strategy. Specifically, we further train the retrieval engine with additional target domain data from Spoken-SAX on top of original BDD-X data samples. As shown in \cref{tab:engine_finetune}, we observe a further performance boost in terms of driving action explanation and justification in the out-of-distribution setting. Although the improvement in performance is less significant in the in-distribution setting, as indicated by the competing justification metric, the action explanation performance has improved considerably. This suggests a higher potential upper bound for retrieval performance and the potential of learning-based optimization for a better retrieval strategy.
}{3.1}


\subsection{Qualitative Demonstration}
We also demonstrate a series of \fixM{qualitative}{2.7.12} examples comparing the driving action explanation and justification provided by human ground truth and prediction from our method. As shown in \cref{fig:demo}, we observe \textit{RAG-Driver} produces robust intelligible action explanation and justification under different environments (i.e. night time and adversarial weather) with a control signal close to the human driver record. More importantly, in the out-of-distribution setting \textit{Spoken-SAX} as indicated by the clear visual difference, we observe the prediction also produces human-understandable answers, qualitatively validating the exceptional zero-shot generalisation capacity.

\section{Limitations and Future Work}
This work aims to develop a generalisable explainable driving agent using a \fixM{Multi-Modal Large}{2.7.51} Language Model (MLLM), addressing a significant obstacle that has hindered deployment: the poor generalisation capacity. \fixM{However, as an early exploration, we still observe several limitations that need to be addressed in future works to realize the application of such promising technology.

\subsection{Computational Restriction and Scale of MLLM}\label{sec:comp}
\label{sec:lim_scale}
While our Multi-modal Large Language Model (MLLM) has shown impressive capabilities in visual reasoning and planning for driving tasks, it is crucial to note that it consists of only $7$ billion parameters. This size is relatively modest compared to more well-known models like GPT-4V \cite{achiam2023gpt} and Gemini \cite{team2023gemini}, which have a significantly larger parameter count and exhibit superb performance in visual understanding and reasoning. Although a prominent trend observed in MLLM work is that performance improvements typically occur in tandem with model scaling, and we expect similar advancements in driving applications, we observe this limitation associated with the current model scale. 

One of the specific limitations is the number of ICL examples that can be used. During training and inference, we provide two ICL examples for each query. This is due to the limitation of the context window size of the LLM backbone, Vicuna 1.5 \cite{zheng2023judging}, which is $4096$ tokens. Each additional ICL example introduces roughly an extra $1800$ tokens, including a $1024$-token fixed-length video sequence and the remaining tokens for language dialogue. While techniques such as Rotary Positional Embeddings (RoPE) \cite{su2024roformer} and slicing windows can help the model to understand longer contexts, the context window -- defining the maximum range of the attention mechanism during a single inference  -- is fundamentally restricted by the model architecture and remains a bottleneck. We expect the advancement of LLM architectures to support larger context window sizes, enabling a more flexible adoption of ICL examples.} 
{1.4.2}

\fixM{\noindent\subsection{Data Scarcity\label{sec:datascarce}} Another challenge for the application of MLLMs in driving is data scarcity. Observing the trend in the development of MLLMs, their success is largely built upon large-scale pre-training and high-quality data for fine-tuning, which enable more powerful and generalisable models. While the integration of language in the decision-making process of an MLLM-based driving agent significantly enhances explainability and provides a human-understandable interface, it introduces a significant paradigm shift from previous deep learning methods, which reveals a lack of paired driving and language datasets. 
Currently, the scale of driving-specific video-caption datasets is considerably smaller than its generic counterparts, setting this as the bottleneck for building robust driving agents. Also, the current driving-language dataset mostly contains driving-explanation pairs (e.g., BDD-X \cite{kim2018bddx}), which are not yet suitable for pre-training in terms of scale and language format. However, we observe more emerging works such as DrivingMLM \cite{wang2023drivemlm} and Reason2Drive \cite{nie2023reason2drive}, which are curating larger scales of well-annotated data specifically for MLLM-based driving applications, providing promising future potential.
}{1.2}



\fixM{
\noindent\subsection{Hallucination\label{sec:hallucinate}} 
MLLMs are known to hallucinate, generating information or responses misaligned with facts \cite{huang2023survey}.
While our retrieval-augmented mechanism for grounding the model's explanations in expert demonstrations has mitigated such phenomena (evident in the much improved Action and Justification explaination metrics in~\cref{tab:benchmark_main}), we have observed some isolated failure cases. For instance, in row 3 column 2 of \cref{fig:demo}, the model associates the car's action of slowing down with the presence of a stop sign, which does not in fact exist. We hypothesise that this is due to the limited capacity of small-scale models which rely on spurious correlations in limited training data. 
We expect a more powerful backbone model and a more balanced dataset to lead to more faithful predictions and explanations.
In the mean time, we would suggest techniques such as real-time verification and rectification \cite{tonmoy2024comprehensive} for mitigating hallucinations in driving applications as a promising direction.}{2.5}

\fixM{\noindent\subsection{Closed-loop Evaluation\label{sec:closedloop}}
Moreover, the close-loop evaluation in simulator is an important step before the deployment of the MLLM-based driving agent.
However, the main obstacle for applying such evaluation in simulator is data scarcity and a significant simulation-to-real gap. 
While works such as DriveMLM \cite{wang2023drivemlm} is able to perform the close-form evaluation by specifically trained on large-scale simulation dataset collected in the same CARLA environment, for methods trained on real dataset like RAG-Driver, the adaptation to simulated environment is a non-trivial step. While our method mitigates the effects of distribution shift between real datasets collected in different location and illumination conditions, the larger simulation-to-real gap still remains a significant challenge for closed-loop testing. This in turn highlights the importance of developing more generalisable models for the deployment of the MLLM-based driving agent.
}{3.2}

\section{Conclusion} 
\label{sec:conclusion}
We propose \textit{RAG-Driver}, a Multi-Modal Large Language Model with Retrieval-augmented In-context Learning capacity designed for generalisable and explainable end-to-end driving. It exhibits strong capability in providing numerical control signals, along with explanations and justifications for driving actions. More importantly, it shows impressive zero-shot generalisation to unseen environments without the need for additional training.

\section*{Acknowledgement}
Bo Zhao was supported by NSFC-62306046.
Lars~Kunze and Daniel Omeiza were supported by EPSRC Project RAILS (EP/W011344/1).
Matthew Gadd was supported by EPSRC Programme Grant ``From Sensing to Collaboration'' (EP/V000748/1).


{\small
\bibliographystyle{plainnat}
\bibliography{main}
}

\clearpage

\appendix

\subsection{Training Details}
\label{apx:training_details}

\noindent\textbf{Embedding Projector}
We leverage a three-layer MLP as the embedding projector to fuse the video $1 \times 1024$ and control signal $1 \times 28$ embedding into a hybrid embedding. The architecture of the lightweight projector is a four-layer MLP with GELU activation. It has an input dimension of $1052$ and an output dimension of $1024$. The margin used in triplet loss is 0.5. We train the model for 200 epochs with a learning rate of $1\mathrm{e}{-5}$ with Adam optimiser. 

\noindent\textbf{MLLM Backbone}
We use a learning rate of $2\mathrm{e}{-5}$ with a cosine scheduler. We use a batch size of $4$ with a gradient accumulation step of $2$ on $8$ \textit{A100 GPUs}, which leads to an effective training batch size of $128$. We use a warm-up strategy in the first 5 epochs with a warm-up ratio of $0.03$. We train the model for $2$ epochs.

\subsection{Baseline Details}

In our comparison, we evaluate several baseline methods. The first, S2VT \cite{venugopalan2015sequence}, utilises an end-to-end sequence-to-sequence model with Long Short-Term Memory (LSTM) networks. It is trained on paired video-sentence data, linking video frame sequences to corresponding word sequences, enabling it to generate descriptive captions for video events. The second method, WAA \cite{kim2018bddx}, employs a visual attention model that trains a convolutional network from images to vehicle control commands. This method focuses on identifying influential image regions through the controller's attention and uses an attention-based video-to-text model to produce textual explanations aligned with the controller's attention maps, grounding the explanations in relevant scene parts. The third approach, ADAPT \cite{jin2023adapt}, is a transformer-based method that leverages a multi-task joint training framework. It aligns driving action captioning with control signal prediction tasks. Finally, DriveGPT4 \cite{xu2023drivegpt4}, trained using LLaVA \cite{liu2023llava} to generate a visual instruction tuning dataset derived from the BDD-X dataset \cite{kim2018bddx}, processes multimodal input data and is capable of generating text responses while predicting control signals, fine-tuned with the assistance of ChatGPT on the new dataset.

\subsection{Linear Layer Parameter Update Derivations}
\label{apx:ICL_Derivation}

Consider a forward-pass through a linear layer $\mathcal{F}$ after its weights $W_0$ have been updated by $\Delta{}W$
$$
\mathbf{y}=\mathcal{F}(\mathbf{x})=(W_0+\Delta{}W)\mathbf{x}
=W_0\mathbf{x}+\Delta{}W\mathbf{x}
$$
The weight update itself is expressed as
$$
\Delta{}W= 
\eta\frac{\partial{}L}{\partial\mathbf{y}}|_{\mathbf{y}_i}\mathbf{x}_i^T
\Rightarrow
\Delta{}W\mathbf{x}= 
\eta\frac{\partial{}L}{\partial\mathbf{y}}|_{\mathbf{y}_i}\mathbf{x}_i^T\mathbf{x}
$$
where $\mathbf{x}_i,\mathbf{y}_i$ are the input and output to the layer that resulted in the weight update.
Now, if we in fact have optimised over a mini-batch of input-outputs  $\mathbf{x}_i,\mathbf{y}_i$ we have
$$
\Delta{}W= 
\sum_i\eta\frac{\partial{}L}{\partial\mathbf{y}}|_{\mathbf{y}_i}\mathbf{x}_i^T
\Rightarrow
\Delta{}W\mathbf{x}= 
\sum_i\eta\frac{\partial{}L}{\partial\mathbf{y}}|_{\mathbf{y}_i}\mathbf{x}_i^T\mathbf{x}
$$
Therefore we have a weighted sum of dot products, which is akin to an attention mechanism.
Indeed, from~\cref{eq:icl_1} we can apply \textit{softmax-free} linear attention expressions such as in~\cite{koohpayegani2024sima} for
$$
\begin{aligned}
& W_V\left[\bm{z}_{icl} ; \bm{z}_{q}\right] \mathrm{softmax}\left(\frac{\left(W_K\bm{z}_{1:n}\right)^T (W_Q\bm{z}_{1:n})}{\sqrt{d^{i}}}\right)\\
& \rightarrow  W_V\left[\bm{z}_{icl} ; \bm{z}_{q}\right]\left(W_K\left[\bm{z}_{icl} ; \bm{z}_{q}\right]\right)^T W_Q\bm{z}_{1:n}
\end{aligned}
$$
Multiplying the linear attention matrices through the stacked in-context and query embeddings we have
$$
W_V\bm{z}_{icl}(W_K\bm{z}_{icl})^TW_Q\bm{z}_{1:n}
+W_V\bm{z}_{q}(W_K\bm{z}_{q})^TW_Q\bm{z}_{1:n}
$$
Now take out a common factor of $W_Q\bm{z}_{1:n}$ for 
$$
(W_V\bm{z}_{icl}(W_K\bm{z}_{icl})^T
+W_V\bm{z}_{q}(W_K\bm{z}_{q})^T)W_Q\bm{z}_{1:n}
$$
and put $\Delta{}W_{ICL}=W_V\bm{z}_{icl}(W_K\bm{z}_{icl})^T$ and $W_{ZSL}=W_V\bm{z}_{q}(W_K\bm{z}_{q})^T$ as the terms both pre-multiplying $W_Q\bm{z}_{1:n}$.
Note that $W_{ZSL}$ is \textit{independent of the in-context terms} (depending only on the query).
Now, in $\Delta{}W_{ICL}$ we in fact have a set of in-context retrieved samples $\bm{z}_{icl,i}$ such that
$$
\begin{aligned}
& W_V\left[\bm{z}_{icl,0},\bm{z}_{icl,1},\ldots\right](W_K\left[\bm{z}_{icl,0},\bm{z}_{icl,1},\ldots\right])^T
\\
& \rightarrow\Delta{}W_{ICL}=\sum_i~W_V\bm{z}_{icl,i}(W_K\bm{z}_{icl,i})^T
\end{aligned}
$$
Finally, by inspection of similar dot-product expressions $W_V\bm{z}_{icl,i}(W_K\bm{z}_{icl,i})^T\leftrightarrow\eta\frac{\partial{}L}{\partial\mathbf{y}}|_{\mathbf{y}_i}\mathbf{x}_i^T\mathbf{x}$ we note that we match the form for the linear layer above
$$
(W_{ZSL}+\Delta{}W_{ICL})W_Q\bm{z}_{1:n}
\leftrightarrow
(W+\Delta{}W)\mathbf{x}
$$
This can therefore be interpreted to say that the output of the attention is adjusted in a meta-optimal way to conform to the samples provided as input context, much like gradient descent on the linear layer would adjust that layer to conform to the mini-batch training data, but \textit{crucially} in the case of \textit{RAG-Driver}, without supervision.

\fixM{\noindent\subsection{Action Representation\label{appx:action}} One potential future research direction is action representation. In our study, we adopted a floating-point number action representation for two main reasons: (1) its validated effectiveness in related research such as robot manipulation (e.g., RT-2 \cite{brohan2023rt}, RT-X \cite{padalkar2023open}) and driving applications (e.g., DriveGPT4 \cite{xu2023drivegpt4}); (2) its intuitive alignment with human understanding. However, general-purpose LLMs are known for their suboptimal performance in predicting floating-point numbers \citep{golkar2023xval}. We hypothesise that this is due to 
the discontinuous encoding of real numbers when represented as language tokens, where such representation does not adhere to the linearity of real number space regarding addition and multiplication.
Although our approach involves fine-tuning the numerical language tokens with a paired visual-language-action dataset, known as symbol tuning \cite{wei2023symbol}, to mitigate this issue, we consider the exploration of alternative methods such as continuous number representation \cite{golkar2023xval} and intermediate controller input \cite{sha2023languagempc} for action representation as a promising avenue for future research.}{1.1}

\end{document}